\documentclass{article}

\usepackage{PRIMEarxiv}

\usepackage[utf8]{inputenc} 
\usepackage[T1]{fontenc}    
\usepackage{hyperref}       
\usepackage{url}            
\usepackage{booktabs}       
\usepackage{amsfonts}       
\usepackage{nicefrac}       
\usepackage{microtype}      
\usepackage{lipsum}
\usepackage{fancyhdr}       
\usepackage{graphicx}       
\usepackage{amsthm}   
\usepackage{amsmath}  
\usepackage{amssymb}  
\usepackage{algorithm}
\usepackage{algpseudocode}
\usepackage{multirow}
\usepackage{amssymb}
\usepackage{authblk}
\usepackage{colortbl}
\usepackage{xcolor}
\usepackage{xurl}

\graphicspath{{media/}}     
\renewcommand{\thefootnote}{\fnsymbol{footnote}}
\theoremstyle{plain}
\theoremstyle{definition} 

\title{OptFormer: Optical Flow-Guided Attention and Phase Space Reconstruction for SST Forecasting}

\author{
        Yin Wang$^{1,5,\dagger,*}$, Chunlin Gong$^{2,\dagger}$, Zhuozhen Xu$^3$, Lehan Zhang$^1$, Xiang Wu$^4$ 
 \\
    \vspace{12pt} 
    \small{$^1$Shandong University of Finance and Economics, Jinan, China} \\
    \small{$^2$University of Minnesota, Minneapolis, United States} \\
    \small{$^3$Wuhan University of Technology, Wuhan, China}\\
    \small{$^4$Anqing Normal University, Anqing, China} \\
    \small{$^5$Jinan Fengdi Intelligent Electronics Co., Ltd, Jinan, China} \\
    \vspace{12pt} 
    \small{wangyin@sdufe.edu.cn, gong0226@umn.edu,zhuozhenxu@whut.edu.cn\\ 202418240230@mail.sdufe.edu.cn, 062313@aqnu.edu.cn}
}

\begin{document}

\maketitle
\begingroup
\renewcommand{\thefootnote}{\fnsymbol{footnote}}
\setcounter{footnote}{1}\footnotetext{Corresponding author.}
\setcounter{footnote}{2}\footnotetext{Equal Contribution.}
\endgroup

\begin{abstract}
Sea Surface Temperature (SST) prediction plays a vital role in climate modeling and disaster forecasting. However, it remains challenging due to its nonlinear spatiotemporal dynamics and extended prediction horizons. To address this, we propose OptFormer, a novel encoder–decoder model that integrates phase-space reconstruction with a motion-aware attention mechanism guided by optical flow. Unlike conventional attention, our approach leverages inter-frame motion cues to highlight relative changes in the spatial field, allowing the model to focus on dynamic regions and capture long-range temporal dependencies more effectively. Experiments on NOAA SST datasets across multiple spatial scales demonstrate that OptFormer achieves superior performance under a 1:1 training-to-prediction setting, significantly outperforming existing baselines in accuracy and robustness.
\end{abstract}


\section{Introduction}
Sea surface temperature (SST) is a key indicator in the global climate system and plays a critical role in understanding air–sea interactions and assessing extreme climate events.
The evolution of SST not only reflects the internal thermal processes of the ocean, but also exerts significant influence on weather systems, energy exchange, and ecological balance \cite{ramanathan1981}. 
Abnormal variations in SST are closely related to tropical cyclones, marine heatwaves, and interannual climate oscillations, which can trigger a wide range of ecological and societal impacts \cite{drayan1983}.
Therefore, improving the modeling and prediction of SST's spatiotemporal dynamics is of fundamental importance to climate science, and also provides crucial support for disaster warning and resource management \cite{xie2004}.

SST prediction methods generally fall into two major categories: physics-based numerical models and data-driven approaches. Numerical models such as ROMS \cite{shchepetkin2005regional} and NEMO \cite{madec2008nemo} simulate ocean processes by explicitly solving dynamical and thermodynamical equations. In comparison, neural network models use data-driven learning to capture spatiotemporal dependencies, with models such as Transformers \cite{vaswani2017attention, zhou2021informer} and ConvLSTM networks \cite{zhou2024spatiotemporal} increasingly applied to SST forecasting tasks.

Despite rapid advances, both categories of SST prediction methods face their limitations. Physics-based numerical models are computationally intensive and sensitive to initial and boundary conditions, limiting their practical scalability. Data-driven approaches often rely on large volumes of labeled data and are difficult to interpret.These limitations point to a challenge: how can we design SST prediction models that are both lightweight and require minimal training data? Can such models be easily deployed on consumer-grade GPUs while still delivering high forecasting accuracy?

Building on this motivation, we propose OptFormer, a lightweight and data-efficient model for SST prediction. OptFormer employs phase-space reconstruction to unfold delayed time series into a high dimensional space, enabling the model to recover the underlying dynamical structure of the SST system. By explicitly modeling the topological relationships between original and delayed attractors, the model gains a deeper understanding of system dynamics, which in turn allows for more effective use of limited training data. In addition, optical flow information is incorporated to enhance the model’s ability to capture spatial dynamics and temporal–frequency patterns essential for accurate SST forecasting. 

The main contributions of this paper are as follows:

\begin{enumerate}
    \item We propose a phase-space reconstruction approach to reconstruct the dynamical structure of SST by unfolding delayed time series in high-dimensional space.
    
    \item We design an encoder–decoder framework with an Optical-Flow-Guided Attention module to enhance spatiotemporal modeling.

    \item We evaluate OptFormer against multiple baselines under single-point and spatiotemporal parallelization, and perform ablation studies to verify the effectiveness of each component.
  \end{enumerate}

The source code and scripts to reproduce the experiments are available at \url{https://anonymous.4open.science/r/OptFormer-Optical-Flow-Guided-Attention-and-Phase-Space-Reconstruction-for-SST-Forecasting-7E1E}.

\section{Related Works}

\subsection{Physics-based models}

Physics-based numerical models simulate SST by solving the governing equations of ocean dynamics using different discretization schemes and coordinate systems.
The Regional Ocean Modeling System (ROMS) \cite{shchepetkin2005regional} uses a split-explicit, free-surface formulation and terrain-following vertical coordinates to solve the primitive equations.
The Nucleus for European Modelling of the Ocean (NEMO) \cite{madec2008nemo} discretizes the ocean using a z-level vertical coordinate system and finite difference methods for large-scale ocean circulation.
The HYbrid Coordinate Ocean Model (HYCOM) \cite{chassignet2007hycom} dynamically switches between isopycnal, terrain-following, and pressure coordinates to better capture stratified ocean layers.
The MIT General Circulation Model (MITgcm) \cite{marshall1997mitgcm} employs finite volume methods on a Cartesian or curvilinear grid and supports both hydrostatic and nonhydrostatic dynamics.

Despite their physical grounding, numerical models are computationally expensive due to the need to solve complex equations over high-resolution grids and long time spans. Their performance is highly sensitive to uncertain initial and boundary conditions, and model tuning often requires substantial expertise and resources, limiting their flexibility for real-time or adaptive forecasting.

\subsection{Data-driven models}
Neural networks based data-driven approaches have become an increasingly important direction for SST prediction.
Zhang et al. \cite{zhang2017prediction} treated SST forecasting as a time series regression problem and used LSTM combined with fully connected layers to support multi-scale prediction with online updating.
Xiao et al. \cite{xiao2019short} developed an ensemble framework combining LSTM with AdaBoost to improve short- and mid-term temperature anomaly forecasting in the East China Sea.
Zhou et al. \cite{zhou2024spatiotemporal} proposed a ConvLSTM-based architecture to jointly model spatial and temporal dependencies in remote sensing data for end-to-end SST prediction.
Zrira et al. \cite{zrira2024time} enhanced bidirectional LSTM with attention mechanisms and applied K-fold cross-validation for robust training on Moroccan coastal SST.
Wang \cite{wang2025stfm_retrieved} incorporated phase-space reconstruction into the STFM framework to better capture the underlying nonlinear dynamics of SST evolution.
Fu et al. \cite{fu2024prediction} integrated LSTM with Transformer modules to exploit both temporal memory and spatial feature extraction in multi-point forecasting.
Yang et al. \cite{yang2024attention-convnet} designed a physics-constrained attention neural network (PANN) by coupling ConvLSTM outputs with PDE dynamics via a cross-attention mechanism.

Despite their flexibility and performance, data-driven models present several key limitations. They often require large volumes of labeled data, which are not always available across regions or seasons. Their black-box nature hinders physical interpretability, and their generalization to out-of-distribution scenarios remains limited. In addition, large models can be computationally demanding, limiting their deployment on resource-constrained platforms.

\section{Methodology}

\subsection{Problem Formulation}

We formulate the evolution of ocean surface temperature as a continuous spatiotemporal dynamical system, discretized for modeling and computation. Let $\Delta t$ be the temporal sampling interval, and define the $i$-th sampling time as:
\begin{equation}
t_i = t_0 + i \cdot \Delta t.
\label{eq:time_sampling}
\end{equation}

At each time $t_i$, the sea surface temperature is measured at $N$ spatial locations, forming a spatial state vector:
\begin{equation}
X(t_i) = [x_1(t_i), x_2(t_i), \ldots, x_N(t_i)] \in \mathbb{R}^N.
\label{eq:spatial_state}
\end{equation}

By collecting $M$ successive observations, we construct the original attractor matrix representing the system's evolution:
\begin{equation}
O = 
\begin{bmatrix}
X(t_1) \\
X(t_2) \\
\vdots \\
X(t_M)
\end{bmatrix}
\in \mathbb{R}^{M \times N}.
\label{eq:spatial_attractor}
\end{equation}

Our goal is to predict the future trajectory of a specific variable $x_k(t)$ over a forecasting horizon of $L$ time steps:
\begin{equation}
\hat{\mathbf{x}}_k^{(L)} = [x_k(t_{M+1}), x_k(t_{M+2}), \ldots, x_k(t_{M+L})].
\label{eq:prediction_target}
\end{equation}

\paragraph{Phase Space Reconstruction.} We assume the temperature field evolves as a deterministic dynamical system with an underlying attractor structure. According to Takens' embedding theorem~\cite{takens2006detecting}, the state of such a system can be reconstructed from time-delayed observations of a single observable. In our context, we aim to reconstruct the local dynamics of $x_k(t)$ using delayed values:
\begin{equation}
Z(t_m) = [x_k(t_m), x_k(t_{m+1}), \ldots, x_k(t_{m+L})],
\label{eq:delay_vector}
\end{equation}
which represents a local trajectory segment starting at time $t_m$. If the original attractor has box-counting dimension $d_o$, and if $L > 2d_o$, then $Z(t_m)$ preserves the topology of the true dynamics~\cite{sauer1991embedology}.

However, we do not directly observe future values of $x_k$ when predicting. Instead, we aim to learn a mapping from the original attractor $O$ to the future trajectory $Z(t_m)$ of a specific location $k$. That is, we seek a function $\Psi$ such that:
\begin{equation}
Z(t_m) = \Psi\left(X(t_m), X(t_{m-1}), \ldots, X(t_{m-M+1})\right).
\label{eq:psi_function}
\end{equation}

According to Eq.\ref{eq:psi_function}, we now construct spatiotemporal translation equation as:
\begin{equation}
\underbrace{
\begin{bmatrix}
x_1(t_1)& x_2(t_1) & \cdots & x_N(t_1) \\
x_1(t_2) & x_2(t_2) & \cdots & x_N(t_2)\\
\vdots &\vdots & \ddots & \vdots \\
x_1(t_M) & x_2(t_M)& \cdots & x_N(t_M)
\end{bmatrix}
}_{\text{Original attractor } O}
\xrightarrow{\Psi}
\underbrace{
\begin{bmatrix}
x_k(t_1) &x_k(t_2) &\cdots & x_k(t_M) \\
x_k(t_2) &x_k(t_3)& \cdots & x_k(t_{M+1}) \\
\vdots &\vdots& \ddots & \vdots \\
x_k(t_L)& x_k(t_{L+1}) & \cdots & x_k(t_{M+L-1})
\end{bmatrix}
}_{\text{Delay attractor } D}
\label{eq:st_mapping}
\end{equation}

This mapping bridges the spatially distributed historical observations and the temporally embedded target prediction. It is the foundation for our learning-based forecasting framework, which models $\Psi$ from data.

\subsection{Network Architecture}

Based on Eq.~\eqref{eq:st_mapping}, we extend the initial attractor $O \in \mathbb{R}^{M \times N}$ and the delay attractor $D \in \mathbb{R}^{L \times M}$ into batch form:  
\begin{equation}
\tilde{O} \in \mathbb{R}^{\text{Batch}\times M\times N}, \quad 
\tilde{D} \in \mathbb{R}^{\text{Batch}\times L\times M}.
\end{equation}
The input observation sequence is $X \in \mathbb{R}^{\text{Batch}\times M\times H\times W}.$ After passing through the Optical Attention module, the optical-flow-related features are extracted, with the spatial dimensions flattened to align with $\tilde{O}$:  
\begin{equation}
\tilde{O} = \text{OpticalAttention}(X) \in \mathbb{R}^{\text{Batch}\times M\times N}, \quad N=H\times W.
\end{equation}  

The current task of the model is then transformed into learning the nonlinear mapping:  
\begin{equation}
\Phi(\tilde{O}) = \tilde{D}.
\end{equation}

Since $\tilde{O}$ has high dimensionality and redundant correlations, directly mapping it into the delay attractor space is inefficient. Therefore, we first apply a linear encoding transformation to compress it into a latent representation $Z$:  
\begin{equation}
Z = W_{e} \cdot \tilde{O} + b_{e}, \quad 
Z \in \mathbb{R}^{\text{Batch}\times M\times d_{\text{model}}},
\end{equation}

where $W_{e}$ and $b_{e}$ represent a set of optimal basis directions learned to compress the redundant correlations within the original attractor point cloud, thereby constructing a compact and structured representation in the latent phase space. The decoder then projects $Z$ into the delay embedding attractor space, achieving nonlinear reconstruction from observations to the phase-space structure:  
\begin{equation}
\tilde{D} = f_{\text{Decoder}}(Z), \quad 
\tilde{D} \in \mathbb{R}^{\text{Batch}\times L\times M}.
\end{equation}

\begin{figure}[htbp]
    \centering
    \includegraphics[width=0.85\linewidth]{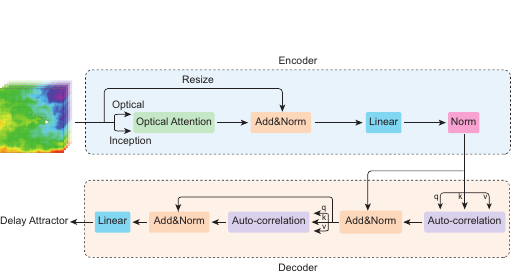}
    \caption{Overall architecture of OptFormer.}
    \label{fig:network}
\end{figure}

Due to the strong periodicity and delayed feedback inherent in ocean–atmosphere interactions~\cite{deser2010sst}, SST time series often exhibit long-range autocorrelated structures. Therefore, We introduce auto-correlation\cite{wu2021autoformer} as a plug-in module to explicitly capture these structured dependencies, offering a natural fit for modeling SST dynamics. The overall structure is shown in Fig.~\ref{fig:network}.  

\subsection{Optical Attention}

Transformer-based models have achieved strong results in science and language domains, yet under small-sample settings, the standard attention mechanism~\cite{vaswani2017attention} tends to overfit and generalize poorly in temporal tasks. To address this, we integrate optical flow~\cite{horn1981determining} into query and key construction, using motion fields as physically grounded attention weights that emphasize high-motion regions. This enhances robustness and temporal awareness with limited data. Additionally, the Inception module~\cite{szegedy2016rethinking} is embedded to capture multi-scale spatial features without downsampling, further strengthening visual representations.

\begin{figure}[htbp]
    \centering
    \includegraphics[width=0.5\linewidth]{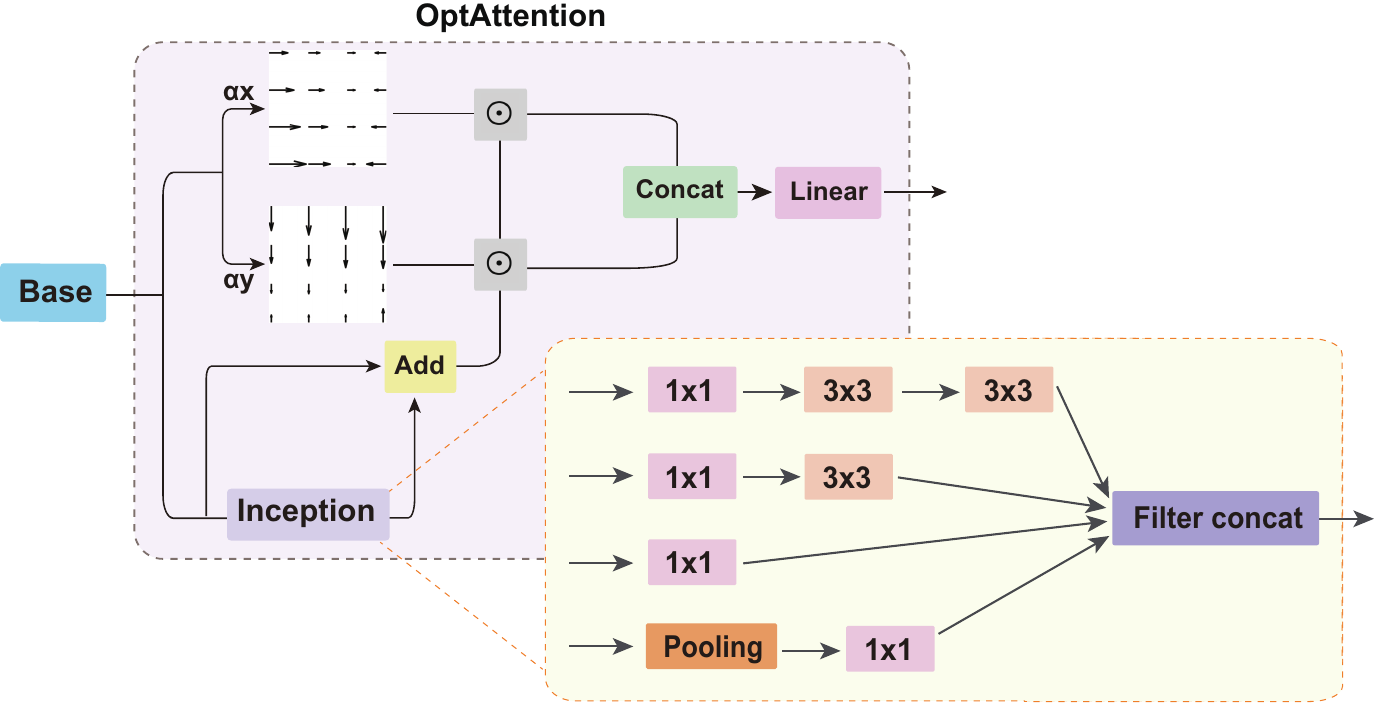}
    \caption{Structure of the Optical Attention module.}
    \label{fig:optical}
\end{figure}

\subsubsection{Optical Flow Estimation}

In this paper, we primarily adopt dense optical flow methods \cite{farneback2003two} for flow estimation. X is a matrix representing discrete samples of the ocean surface temperature at a given moment, by using a binomial expansion, for any point in the ocean surface temperature $X(i,j)$ can be approximated in the local neighborhood as:  
\begin{equation}
X(i,j)\approx \mathbf{x}^{T}A\mathbf{x}+\mathbf{b}^{T}\mathbf{x}+c,
\end{equation}

where $\mathbf{x}=[i-i_{0},j-j_{0}]^{T}$ is the local coordinate relative to the center pixel $(i_{0},j_{0})$, representing the strongly correlated sea areas surrounding a specific point in the sea. $A \in \mathbb{R}^{2\times 2}$ is a symmetric matrix, $\mathbf{b}\in \mathbb{R}^{2}$ is a linear coefficient, and $c \in \mathbb{R}$ is a constant term.  

Using Gaussian-weighted least squares fitting, for two consecutive frames $X_1$ and $X_2$:  
\begin{equation}
X_{1}(\mathbf{x}) = \mathbf{x}^{T}A_{1}\mathbf{x}+\mathbf{b}_{1}^{T}\mathbf{x}+c_{1}, \quad 
X_{2}(\mathbf{x}) = \mathbf{x}^{T}A_{2}\mathbf{x}+\mathbf{b}_{2}^{T}\mathbf{x}+c_{2}.
\end{equation}

Based on the brightness constancy assumption $X_{1}(i,j)=X_{2}(i+u,j+v)$, the displacement vector $\mathbf{d}=[u,v]^{T}$ satisfies:  
\begin{equation}
X_{2}(\mathbf{x}+\mathbf{d}) \approx (\mathbf{x}+\mathbf{d})^{T}A_{2}(\mathbf{x}+\mathbf{d})+\mathbf{b}_{2}^{T}(\mathbf{x}+\mathbf{d})+c_{2}.
\end{equation}  

For small displacements, the approximate intensity difference is:  
\begin{equation}
\Delta X(\mathbf{x}) = X_{2}(\mathbf{x}) - X_{1}(\mathbf{x}) \approx \mathbf{b}_{1}^{T}\mathbf{d}.
\end{equation}

The displacement is optimized by minimizing the data term within the window:  
\begin{equation}
E(d)=\sum_{\mathbf{x}\in \Omega}{w(\mathbf{x})\left( \mathbf{b}_{1}^{T}\mathbf{d}(\mathbf{x})-\Delta X(\mathbf{x}) \right)^{2}},
\end{equation}  
where $w(x)$ is the Gaussian weight, and $\Omega$ denotes the local neighborhood.  

To capture large displacements, a Gaussian pyramid is employed. At level $l$, frames are downsampled as:  
\begin{equation}
X^{(l)}(i,j) = X^{(l-1)}\left(\frac{i}{2},\frac{j}{2}\right), \quad l=0,\ldots,3.
\end{equation}  

The displacement is updated iteratively from coarse to fine scales:  
\begin{equation}
\mathbf{d}^{(l-1)}(i,j) = 2\cdot \mathbf{d}^{(l)}\left(\frac{i}{2},\frac{j}{2}\right)+\Delta \mathbf{d}^{(l-1)}.
\end{equation}   

Finally, to ensure spatial continuity of the optical flow field, a regularization term is introduced to optimize the total energy:  
\begin{equation}
E_{\text{total}}(\mathbf{d})=\sum_{\mathbf{x}}w(\mathbf{x})\left( \mathbf{b}_{1}^{T}\mathbf{d}(\mathbf{x})-\Delta X(\mathbf{x}) \right)^{2}+\lambda \sum_{\mathbf{x}} ||\nabla \mathbf{d}(\mathbf{x})||^{2},
\end{equation}
where $\nabla \mathbf{d}=[\partial u/\partial i, \partial u/\partial j, \partial v/\partial i, \partial v/\partial j]^{T}$, and $\lambda$ balances data consistency and smoothness. Gaussian filtering is used to enhance robustness.  

We compute optical flows for $M-1$ consecutive frame pairs $m\in \{0,...,M-2\}$. To align with the $M$ observation frames, an additional flow is extracted from the final frame pair, yielding the $M$-th optical flow field:  

\begin{equation}
\begin{aligned}
\text{flow}_{\text{data}}[m,:,:,:] &= \mathbf{d}(i,j) = [u(i,j), v(i,j)], \quad m = 0, \ldots, M-1, \\
\alpha_{x} &= \text{flow}_{\text{data}}[:,:,:,0], \quad 
\alpha_{y} = \text{flow}_{\text{data}}[:,:,:,1].
\end{aligned}
\label{eq:optical_flow}
\end{equation}

\begin{figure}[htbp]
    \centering
    \includegraphics[width=0.5\linewidth]{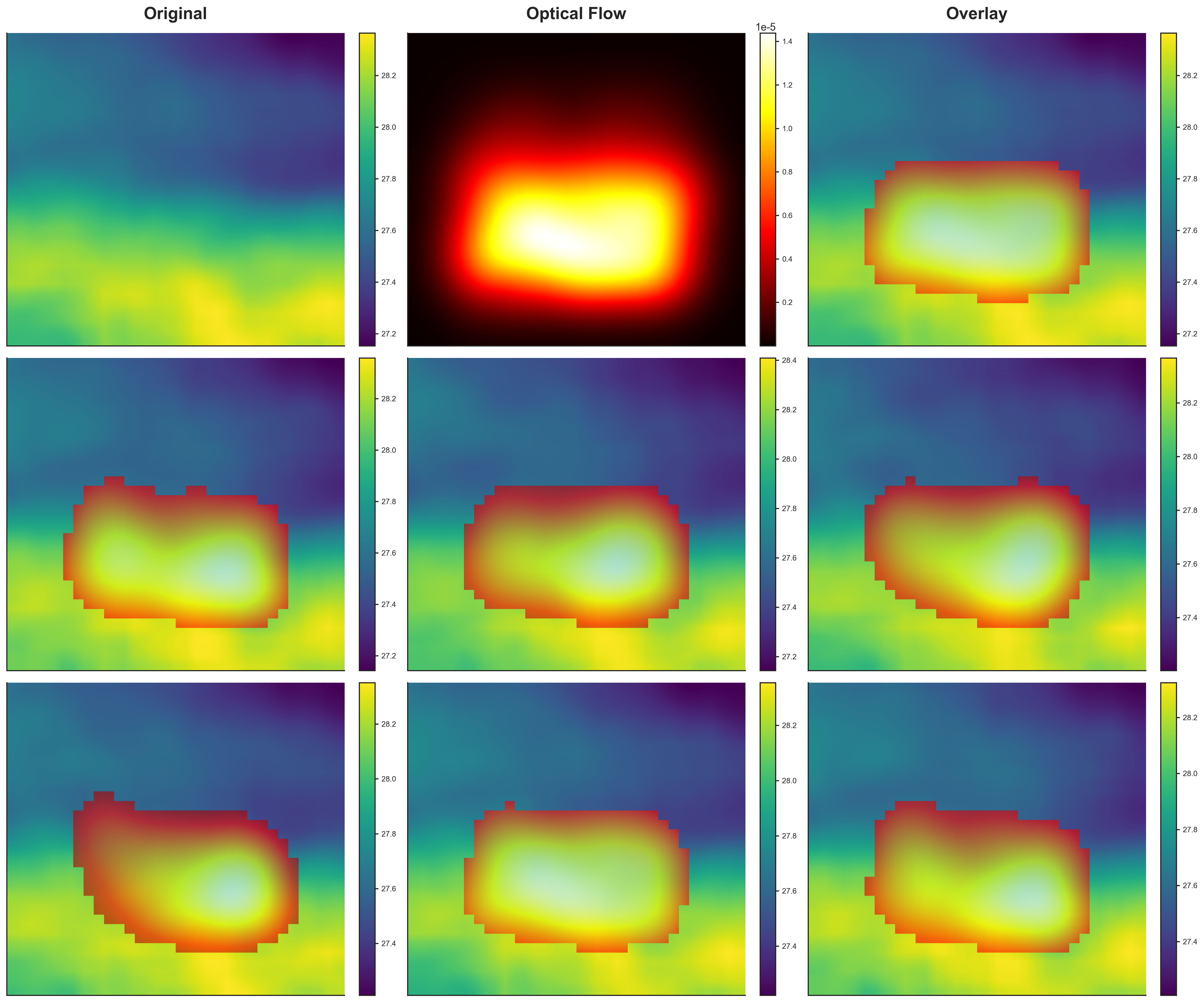}
    \caption{Temporal evolution of SST data with optical flow overlay. The first row displays the original SST field, optical flow magnitude, and their overlay at the initial time step (t=0). The second and third rows show the evolution of the overlay at subsequent time steps (t=1-6).}
    \label{fig:flow}
\end{figure}

Here, $\alpha_{x}$ represents the optical flow vectors in the X-direction, and $\alpha_{y}$ those in the Y-direction. These two directional flows capture the dynamic variations in the observed data. The visualized optical flow vector map is shown in Fig.~\ref{fig:flow}, which provides an intuitive representation of motion distribution across spatial directions.       

\subsubsection{Information Aggregation}

We first extract optical flow vectors $\alpha_{x},\alpha_{y}$ from the raw observation sampling $X$ of size $Batch\times M\times H\times W$, which are used to guide spatial attention. Meanwhile, the raw observation $X$ is fed into the Inception module to obtain a stable-dimensional deep feature representation matrix $V$:
\begin{equation}
V=(Inception(X)+X).reshape(Batch,M,N),
\end{equation}
thereby enhancing the expressive ability of the ocean observation data.  

The Inception module captures both local and global features by combining convolutions of different scales in parallel paths, resulting in richer and more robust representations.  

The optical flow vectors $\alpha_{x},\alpha_{y}$ are then fused with the image features $V$:  
\begin{equation}
\begin{aligned}
X &= \alpha_{x}.\text{reshape}(\text{Batch} \times M \times N) \odot V, \\
Y &= \alpha_{y}.\text{reshape}(\text{Batch} \times M \times N) \odot V.
\end{aligned}
\label{eq:optical_flow_weighting}
\end{equation}

We concatenate the fused optical flow vectors in the X and Y directions along the last dimension, and then apply a linear layer to compress the feature dimensions:  
\begin{equation}
Integral = concat(X,Y,Axis=-1), \quad 
\tilde{O}  = Linear(Integral).
\end{equation}

\section{Experiment Design}

\subsection{Dataset}
All experiments are conducted on the NOAA Optimum Interpolation Sea Surface Temperature (OISST v2.1) dataset\cite{huang2021doisst}.
This dataset provides daily SST observations with a spatial resolution of $0.25^\circ \times 0.25^\circ$, enabling detailed characterization of spatiotemporal SST variations. 
The experimental region is selected from the central Atlantic Ocean, where ocean currents are active and air--sea interactions are pronounced, exhibiting typical nonlinear dynamical behaviors. 
Such characteristics make this region well suited for evaluating predictive models designed for complex marine systems. 
Unless otherwise specified, the dataset is split into training and testing subsets with a ratio of $1{:}1$ along the temporal dimension.

\subsection{Data Construction}
To model spatial dependencies and temporal dynamics in the ocean system, we adopt a data construction strategy that combines regional association with a sliding window strategy. 
In the spatial domain, state association units are constructed using fixed-size latitude--longitude windows to capture potential dynamical coupling among neighboring ocean regions. 
In the temporal domain, time-series samples are generated using a sliding window, and delay-embedding techniques are applied to reconstruct the underlying dynamical attractor of the system. 
This construction allows the model to effectively learn complex spatiotemporal dynamics from limited observations. 

\subsection{Baselines}
To comprehensively evaluate the predictive performance of OptFormer, several representative time-series forecasting models are selected as baselines, including LSTM\cite{hochreiter1997long}, Informer\cite{zhou2021informer}, AutoFormer\cite{wu2021autoformer}, N-BEATS\cite{oreshkin2019n-beats}, TCN\cite{lea2016temporal}.

\subsection{Implementation Details}

\subsubsection{Hyperparameter Settings}
\begin{figure}[htbp]
    \centering
    \includegraphics[width=0.4\linewidth]{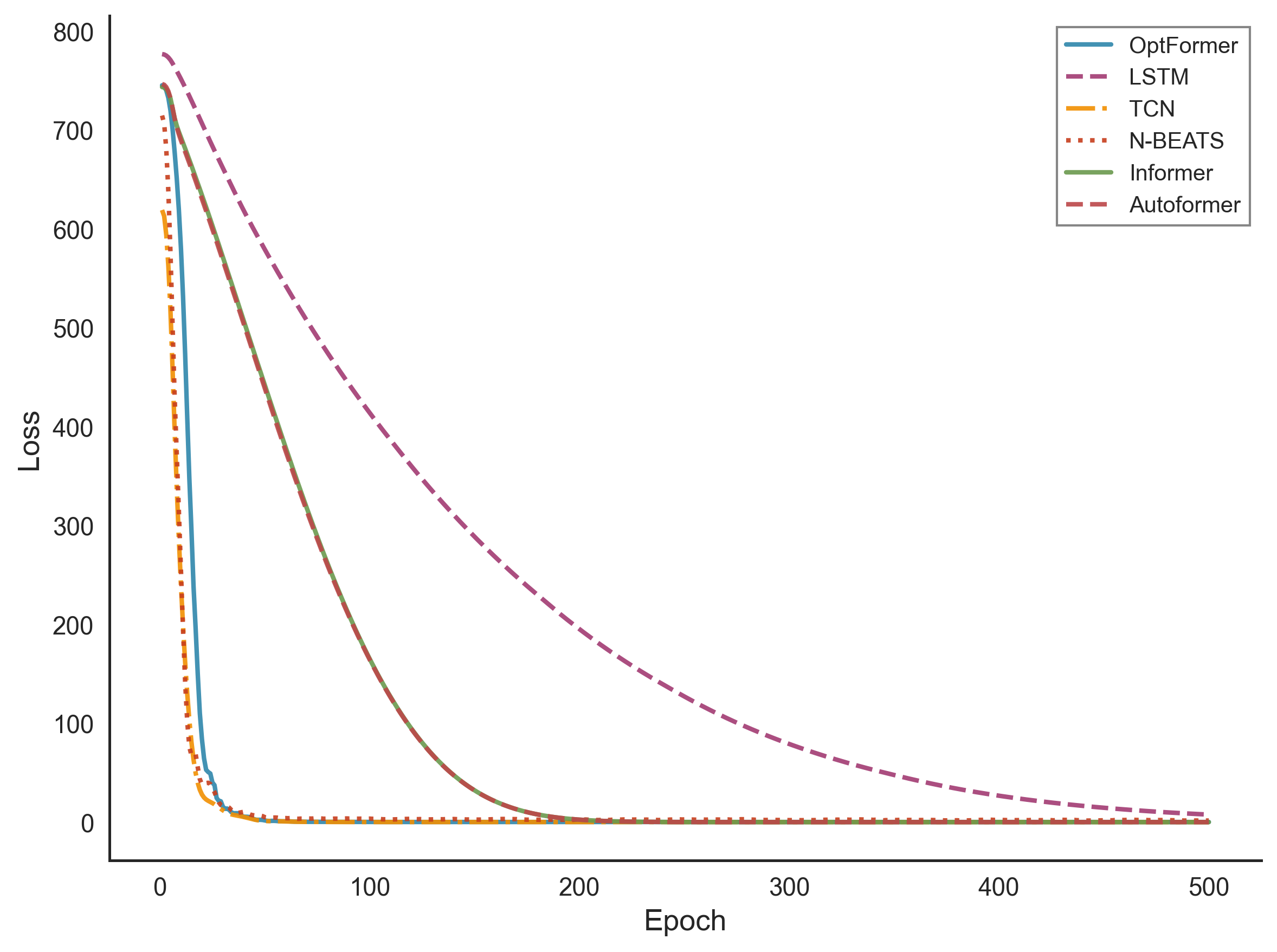}
    \caption{All models are evaluated at their respective converged epochs to ensure a fair comparison. As observed, OptFormer converges significantly faster than the baseline methods, indicating more efficient optimization and faster fitting to the underlying spatiotemporal dynamics.}
    \label{Loss}
\end{figure}

Under the default configuration, the hidden dimension of OptFormer is set to $128$, and the hidden dimension of the feed-forward network is set to $256$. 
Time-series samples are constructed using a sliding window of length $30$, with the training and prediction horizons kept identical to ensure sufficient historical information for delay-embedded attractor reconstruction. 
The temporal gap between adjacent windows is fixed to $t_{\text{gap}} = 5$. 
For fair comparison, as shown in Fig.\ref{Loss}, all baseline models are evaluated using their respective converged training epochs under the same data settings.

\subsubsection{Training Settings}
All models are trained on a single NVIDIA GeForce RTX 3060 (12GB) GPU. 
Under the default configuration, the whole training requires approximately $10$ seconds. 
The batch size is fixed to $30$, and the total number of training epochs is set to $220$. 
All experiments follow identical training settings to ensure the consistency and fairness of performance comparisons.

\subsubsection{Evaluation Metrics}
Prediction performance is evaluated using the Root Mean Square Error (RMSE) and the Mean Absolute Percentage Error (MAPE). 
RMSE measures the overall prediction accuracy, while MAPE quantifies the relative error level. 
For experiments involving multiple parameter settings or scenarios, the final results are reported as averaged errors to provide stable and representative performance estimates.

\section{Result Analysis}
\subsection{Parameter Experiment}

To evaluate the impact of spatial area dependency on the model's prediction performance, this study constructs experimental sub-regions with different longitude-latitude ranges and compares their prediction accuracy differences. As shown in Tab\ref{tab:1}, the model's root mean square error (RMSE) generally decreases as the spatial coverage expands, indicating that a larger spatial domain can provide richer system state information, which helps capture key dynamic coupling structures, thereby improving the reconstruction quality and prediction accuracy of the initial attractor. The results show that OptFormer exhibits significantly superior prediction performance across all spatial scales and maintains high stability within different area ranges.
\begin{table}[h]
\centering
\small
\begin{tabular}{l
cc @{\hspace{6pt}}
cc @{\hspace{6pt}}
cc}
\toprule
\textbf{Model} 
& \multicolumn{2}{c}{\textbf{1°$\times$1°}} 
& \multicolumn{2}{c}{\textbf{2°$\times$2°}} 
& \multicolumn{2}{c}{\textbf{4°$\times$4°}} \\
\cmidrule(lr){2-3}
\cmidrule(lr){4-5}
\cmidrule(lr){6-7}
& RMSE & MAPE 
& RMSE & MAPE 
& RMSE & MAPE \\
\midrule
OptFormer 
& \textbf{0.485} & \textbf{1.600}\% 
& \textbf{0.355} & \textbf{1.104}\% 
& \textbf{0.373} & \textbf{1.174}\% \\

LSTM 
& 1.092 & 3.883\% 
& 1.098 & 3.848\% 
& 1.075 & 3.821\% \\

AutoFormer 
& 1.106 & 3.930\% 
& 1.106 & 3.905\% 
& 1.101 & 3.913\% \\

N-BEATS 
& 2.911 & 10.477\% 
& 2.711 & 9.735\% 
& 2.038 & 7.254\% \\

TCN 
& 1.248 & 4.328\% 
& 1.179 & 4.138\% 
& 1.065 & 3.670\% \\

Informer 
& 1.163 & 4.151\% 
& 1.157 & 4.132\% 
& 1.100 & 3.914\% \\

\bottomrule
\end{tabular}
\caption{Comparison of different models under different spatial sampling areas. Best results for each resolution and metric are highlighted in bold.}
\label{tab:1}
\end{table}

As shown in Tab.\ref{tab:2}, OptFormer achieves significantly superior prediction performance across different sampling intervals ($\Delta t = 1, 3, 5$). Compared to the model with the highest error, OptFormer achieves an average relative improvement of over 76\% in RMSE and an average relative decrease of over 78\% in MAPE, fully demonstrating its high robustness and accurate modeling ability against changes in time scale.

\begin{table}[htbp]
\centering
\small
\begin{tabular}{l
cc @{\hspace{6pt}}
cc @{\hspace{6pt}}
cc}
\toprule
\textbf{Model} 
& \multicolumn{2}{c}{$\boldsymbol{\Delta t = 1}$} 
& \multicolumn{2}{c}{$\boldsymbol{\Delta t = 3}$} 
& \multicolumn{2}{c}{$\boldsymbol{\Delta t = 5}$} \\
\cmidrule(lr){2-3}
\cmidrule(lr){4-5}
\cmidrule(lr){6-7}
& RMSE & MAPE 
& RMSE & MAPE 
& RMSE & MAPE \\
\midrule
OptFormer 
& \textbf{0.446} & 1.403\% 
& \textbf{0.868} & \textbf{2.971}\% 
& \textbf{0.373} & \textbf{1.174}\% \\

LSTM 
& 0.467 & 1.481\% 
& 1.144 & 4.103\% 
& 1.096 & 3.901\% \\

AutoFormer 
& 0.547 & 1.743\% 
& 1.055 & 3.740\% 
& 0.977 & 3.443\% \\

N-BEATS 
& 3.067 & 11.063\% 
& 3.451 & 12.454\% 
& 2.038 & 7.254\% \\

TCN 
& 0.682 & 2.242\% 
& 1.822 & 6.572\% 
& 1.135 & 3.945\% \\

Informer 
& 1.597 & 5.679\% 
& 0.988 & 3.328\% 
& 1.156 & 4.039\% \\

\bottomrule
\end{tabular}
\caption{Comparison of different models under different time delays. Best results for each time delay and metric are highlighted in bold.}
\label{tab:2}
\end{table}

Furthermore, to evaluate the model's performance at different prediction lengths, we tested it in short-term ($L=15$), mid-term ($L=30$), and long-term ($L=45$) in Tab.\ref{tab:3}. The experimental results show that when the prediction length is short ($L=15$), there is a certain error in the construction of the initial attractor due to insufficient historical information, leading to a decrease in prediction accuracy. In the longer prediction period ($L=45$), the system attractor may undergo a phased shift, making it difficult for the model to accurately reconstruct both the initial and delayed attractors simultaneously, thus affecting overall performance. In contrast, when $L=30$, the model achieves the optimal prediction effect, indicating that this duration strikes a good balance between information sufficiency and attractor stability.
\begin{table}[htbp]
\centering
\small
\begin{tabular}{l
cc @{\hspace{6pt}}
cc @{\hspace{6pt}}
cc}
\toprule
\textbf{Model} 
& \multicolumn{2}{c}{\textbf{L = 15}} 
& \multicolumn{2}{c}{\textbf{L = 30}} 
& \multicolumn{2}{c}{\textbf{L = 45}} \\
\cmidrule(lr){2-3}
\cmidrule(lr){4-5}
\cmidrule(lr){6-7}
& RMSE & MAPE 
& RMSE & MAPE 
& RMSE & MAPE \\
\midrule
OptFormer 
& \textbf{0.609} & \textbf{2.012}\% 
& \textbf{0.351} & \textbf{1.102}\% 
& \textbf{0.739} & \textbf{2.356}\% \\

LSTM 
& 1.153 & 4.192\% 
& 1.091 & 3.881\% 
& 1.361 & 4.675\% \\

AutoFormer 
& 1.175 & 4.278\% 
& 1.098 & 3.905\% 
& 1.308 & 4.494\% \\

N-BEATS 
& 0.927 & 3.214\% 
& 1.926 & 6.857\% 
& 3.962 & 13.997\% \\

TCN 
& 2.175 & 8.138\% 
& 1.166 & 4.043\% 
& 1.720 & 5.918\% \\

Informer 
& 1.218 & 4.453\% 
& 1.109 & 3.939\% 
& 1.311 & 4.501\% \\

\bottomrule
\end{tabular}
\caption{Comparison of different models under different prediction lengths. Best results for each prediction length and metric are highlighted in bold.}
\label{tab:3}
\end{table}

\subsection{Parallelization Experiment}
To reduce the evaluation bias caused by relying on a single spatial sample---such as sampling randomness and regional specificity---we propose a spatially parallel sliding-sampling strategy that enhances the representativeness and robustness of model performance across diverse local environments.
This strategy applies a $2^\circ \times 2^\circ$ fixed-size sliding window across a $1^\circ \times 1^\circ$ evaluation space, centering the window on each grid point to extract local subsets for delay-coordinate attractor reconstruction and prediction, with final performance metrics averaged over all windows to yield a stable global accuracy estimate.

To evaluate temporal generalization, we conducted seasonal experiments using four different start times corresponding to spring, summer, autumn, and winter. As shown in Tab.\ref{tab:4}, OptFormer consistently outperforms all baselines across seasons, demonstrating robust adaptability to varying temporal dynamics. Notably, it achieves the lowest error in autumn (RMSE = 0.368), where spatiotemporal patterns are often more regular due to transitional ocean–atmosphere conditions. Compared to the next-best method, OptFormer reduces RMSE by an average of over 40.6\% across all seasons, confirming its strong generalization and long-horizon forecasting capability.

\begin{table}[htbp]
\centering

\begin{tabular}{l
cc @{\hspace{6pt}}
cc @{\hspace{6pt}}
cc @{\hspace{6pt}}
cc}
\toprule
\textbf{Model} 
& \multicolumn{2}{c}{\textbf{Spring}} 
& \multicolumn{2}{c}{\textbf{Summer}} 
& \multicolumn{2}{c}{\textbf{Autumn}} 
& \multicolumn{2}{c}{\textbf{Winter}} \\
\cmidrule(lr){2-3}
\cmidrule(lr){4-5}
\cmidrule(lr){6-7}
\cmidrule(lr){8-9}
& RMSE & MAPE 
& RMSE & MAPE 
& RMSE & MAPE 
& RMSE & MAPE \\
\midrule
OptFormer 
& \textbf{0.763} & \textbf{2.603}\% 
& \textbf{0.705} & \textbf{2.395}\% 
& \textbf{0.368} & \textbf{1.157}\% 
& \textbf{0.848} & \textbf{2.862}\% \\

LSTM 
& 1.210 & 4.187\% 
& 1.207 & 4.177\% 
& 1.083 & 3.853\% 
& 1.130 & 3.833\% \\

AutoFormer 
& 1.194 & 4.136\% 
& 1.198 & 4.151\% 
& 1.103 & 3.923\% 
& 1.097 & 3.709\% \\

N-BEATS 
& 3.256 & 11.526\% 
& 3.459 & 12.263\% 
& 3.302 & 11.905\% 
& 3.243 & 11.536\% \\

TCN 
& 1.191 & 4.066\% 
& 1.461 & 4.960\% 
& 1.003 & 3.432\% 
& 2.349 & 8.294\% \\

Informer 
& 1.192 & 4.131\% 
& 1.188 & 4.121\% 
& 1.111 & 3.956\% 
& 1.102 & 3.726\% \\

\toprule
\rowcolor{gray!15}
Average Change 
& +110.8\% & +115.5\% 
& +141.5\% & +147.8\% 
& +313.2\% & +367.9\% 
& +110.4\% & +117.3\% \\
\bottomrule
\end{tabular}
\caption{Comparison of different models across seasons (RMSE and MAPE). Average relative changes are computed with respect to OptFormer (lower is better). Best results for each season and metric are highlighted in bold.}
\label{tab:4}
\end{table}

\subsection{Ablation Experiment}

We conducted ablation studies to evaluate the role of each component in OptFormer. As shown in Tab.~\ref{tab:6}, removing any module leads to performance degradation, confirming their complementarity. The largest drop occurs when Optical Attention is removed (RMSE +0.079), highlighting its importance in capturing motion-sensitive spatial dynamics. Inception contributes multi-scale spatial encoding (RMSE +0.055), while Auto-Correlation models periodic temporal dependencies (RMSE +0.033), further enhancing long-range prediction.

\begin{table}[H]
\centering

\begin{tabular}{lcccc}
\toprule
\textbf{Component} & \multicolumn{4}{c}{\textbf{Choice}} \\
\midrule
Optical\_Attention & $\checkmark$ & $\times$ & $\checkmark$ & $\checkmark$ \\
Inception          & $\checkmark$ & $\checkmark$ & $\times$ & $\checkmark$ \\
AutoCorrelation    & $\checkmark$ & $\checkmark$ & $\checkmark$ & $\times$ \\
\midrule
RMSE & 0.3558 & 0.4348 & 0.4108 & 0.3898 \\
MAPE & 1.1094\% & 1.3953\% & 1.3052\% & 1.2520\% \\
\bottomrule
\end{tabular}
\caption{Model Performance with Different Components}
\label{tab:6}
\end{table}

\section{Conclusion}

This paper proposes the OptFormer model, which combines phase space reconstruction with an optical flow-guided mechanism, effectively improving the accuracy and robustness of spatiotemporal prediction of oceanic multivariate data. Experimental results across multiple scenarios demonstrate that, in terms of indicators such as RMSE, OptFormer significantly outperforms mainstream methods, achieving both high predictive accuracy and cross-temporal generalization ability. This provides a data-driven approach with both theoretical depth and practical value for ocean environment modeling.

Despite achieving superior results, the model still has room for improvement. At present, only single-variable SST data are used, without considering external forcing factors such as wind stress and geostrophic current, which limits the characterization of complex coupled processes. In the future, multi-source meteorological and ocean observation data will be introduced to build a multivariate joint modeling framework. Furthermore, combined with related studies \cite{long2018pde-net,gong2025afd-sta}, the application potential and generalization ability of OptFormer in typical nonlinear spatiotemporal systems such as partial differential equations will be further explored.

\section*{Acknowledgments}
This paper was supported by the Shandong Provincial Natural Science Foundation (Grant No. ZR2024QD287). We acknowledge NOAA/NCEI for providing the OISST v2.1 SST dataset\cite{huang2020oisst_dataset}.

\end{document}